%% file: main.tex
\documentclass[a4paper,10pt,twocolumn,3p]{elsarticle}

\usepackage[utf8]{inputenc}
\usepackage{lineno,hyperref}
\usepackage{xcolor}
\usepackage{multirow}

\usepackage{comment}

\usepackage{amsmath}

\modulolinenumbers[5]

\usepackage{caption}
\usepackage{subcaption}

\usepackage{url}

\usepackage{breakurl}

\makeatletter
\def\ps@pprintTitle{%
  \let\@oddhead\@empty
  \let\@evenhead\@empty
  \let\@oddfoot\@empty
  \let\@evenfoot\@oddfoot
}
\makeatother

\begin{document}

\begin{frontmatter}

\title{Image-Based Detection of Modifications in PCBs with Deep Convolutional Autoencoders}

\author[addr]{Diulhio Candido de Oliveira\corref{mycorrespondingauthor}}
\cortext[mycorrespondingauthor]{Corresponding author}
\ead{diulhio@alunos.utfpr.edu.br}

\author[addr]{Bogdan Tomoyuki Nassu}
\ead{btnassu@utfpr.edu.br}

\author[addr,addr2]{Marco Aurelio Wehrmeister}
\ead{wehrmeister@utfpr.edu.br}

\address[addr]{Departament of Informatics, Federal University of Technology - Parana (UTFPR), 
80230-901 Curitiba, Brazil}

\address[addr2]{Computer Science Department, University of Münster, 48149 Münster, Germany}

\begin{abstract}
In this paper, we introduce a semi-supervised approach for detecting modifications in assembled printed circuit boards (PCBs) based on photographs taken without tight control over perspective and illumination conditions.
We take as an instance of this problem the visual inspection of gas pump PCBs, which can be modified by fraudsters wishing to deceive customers or evade taxes.
Given the uncontrolled environment and the huge number of possible modifications, we address the problem as a case of anomaly detection, proposing an approach that is directed towards the characteristics of that scenario, while being well-suited for other similar applications.
We propose a loss function that can be used to train a deep convolutional autoencoder based only on images of the unmodified board --- that allows overcoming the challenge of producing a representative set of samples containing anomalies for supervised learning.
We also propose a function that explores higher level features for comparing the input image and the reconstruction produced by the autoencoder, allowing the segmentation of structures and components that differ between them.
Experiments performed on a dataset built to represent real-world situations (and which we will make publicly available) show that our approach outperforms other state-of-the-art approaches for anomaly segmentation in the considered scenario, while producing comparable results on a more general object anomaly detection task.
\end{abstract}


\end{frontmatter}


\input{1_introduction}

\input{2_related_work}

\input{4_method}

\input{5_experiments}

\input{6_conclusion}

\bibliography{ref_paper1}

\end{document}

%% file: 1_introduction.tex
\section{Introduction}

\begin{figure}[h]
    \includegraphics[width=\linewidth]{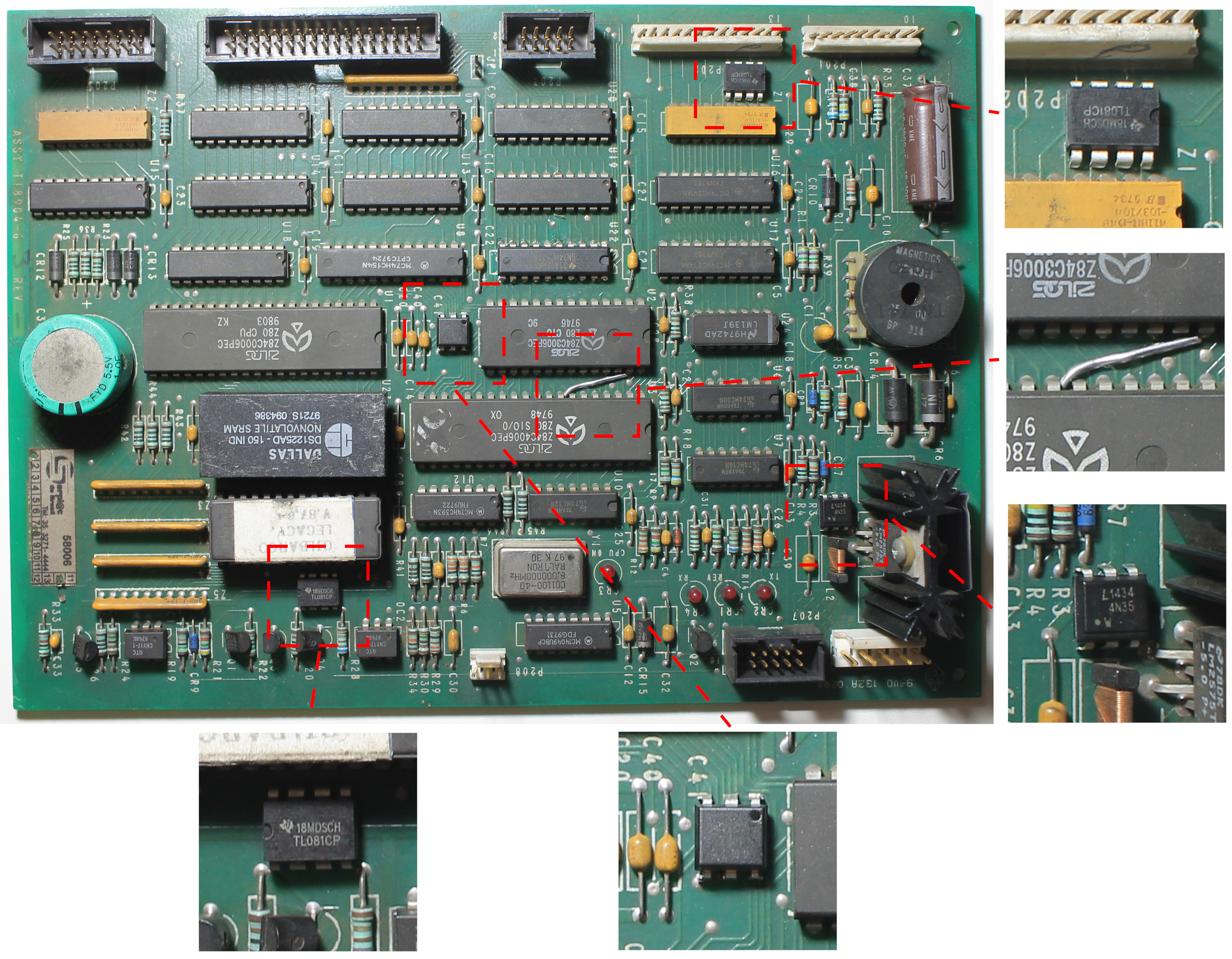}
    \caption {An example of PCB containing modifications. Some of them are easier to identify, while others require more attention. The modifications consist of adding small IC chips, and a jump wire in one case.}
    \label {fig:img_anomalies}
\end{figure}

Detecting anomalies in assembled printed circuit boards (PCBs) is an important problem for fields such as quality control in manufacturing \citep{Bergmann, Cohen2020} and fraud detection \citep{Oliveira2017}.
One instance of the latter is the detection of frauds in gas pumps, a common problem in countries such as Brazil and India \citep{Chakraborty2017, Slattery2021}.
For example, modifying the gas pump PCB by replacing, adding, or removing components allows offenders to force the pump to display a fuel volume different from the one actually put into the tank.
It may be difficult for law enforcers to detect this kind of fraud simply by testing the pump, since the offender can use a remote control to deactivate the fraud during inspections.
Thus, inspectors have to remove the PCB from the gas pump and visually compare the suspicious board to a reference design or sample --- for example, in Brazil gas pump PCB designs are approved and controlled by a regulatory body, and cannot be changed without authorization.
To avoid worries such as legal action from gas station owners who lose profits while the pump cannot be operated, inspections should be quick, but this is frequently not possible, given the complexity of these PCBs.
Figure~\ref{fig:img_anomalies} shows an example of a PCB containing modifications --- the amount of small components makes it hard even for a specialist to notice these modifications.
The task is further complicated if inspectors are not specialists, which leads them to rely solely on visual comparisons.
For these reasons, a system that assists inspectors by automatically detecting modifications or suspicious regions can be an interesting proposition.
Such a system must be flexible enough to work on-site, without requiring large capture structures, controlled lighting or fixed camera positioning.

While we have the fraud detection scenario as our main motivation, this problem shares most characteristics with image-based inspection of PCBs in general --- a task for which several methods have been proposed in recent years.
Some methods are used to detect defects in unassembled PCBs \citep{Adibhatla2020, Shi2020}, where common anomalies are missing holes and open circuits, while other methods deal with assembled PCBs \citep{Oliveira2017, Li2020, MallaiyanSathiaseelan2021}.
These methods are usually based on supervised machine learning, where a decision model is trained by observing samples both with and without defects or anomalies.
One of the foremost challenges when working with this kind of data-driven technique is providing a representative dataset containing a wide range of situations that reflect the actual variety of possibilities faced in practice well enough to allow generalization.
For an unmodified board, that means having samples with varied lighting conditions and camera angles, but anomaly samples are harder to obtain, because they are rare, expensive to reproduce, or may manifest in unpredictable ways.

We adopt a semi-supervised approach, and address the task as an anomaly detection problem.
In this formulation, models are trained only on normal samples, learning to describe their distribution, based on the premise that it is possible to detect anomalies based on how well the learned model is able to describe a given sample --- i.e.~samples containing anomalies are not well described by the model, and will appear as outliers.
Many recent studies aimed at industrial inspection in various settings explore this idea \citep{Bergmann, Cohen2020, Defard2021, Bergmann2021, Shi2021, Wang2021}.

\begin{figure*}[h]
    \includegraphics[width=0.7\linewidth]{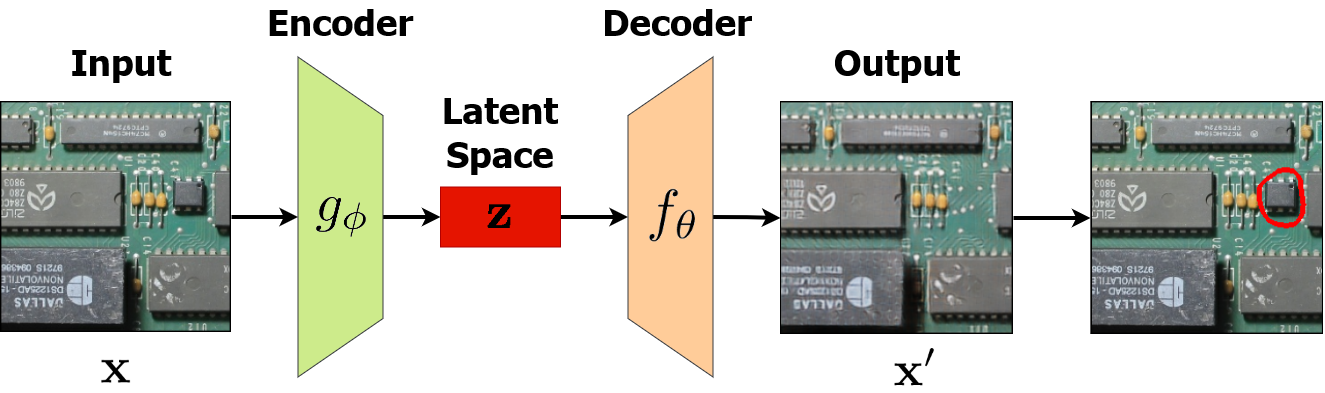}
    \centering
    \caption {The reconstruction-based inference process using a convolutional autoencoder. The autoencoder is trained to reconstruct only anomaly-free samples, so when it receives as input an image containing anomalies, the reconstructed output does not show the modification. Thus, it is possible to segment the anomaly by comparing the input and the output.}
    \label {fig:basic_autoencoder}
\end{figure*}

In this paper, we address the problem of detecting modifications in a PCB using a deep neural network.
More specifically, we propose using a convolutional autoencoder architecture for reconstruction-based anomaly detection.
This kind of architecture compresses the input image to a feature vector, called ``latent space'', and then reconstructs the same image based only on these features.
The rationale behind the proposed method is that if the model is trained only with anomaly-free samples, it will be able to reconstruct only this kind of sample.
Thus, when it receives an image containing anomalies as input, it will be unable to properly reconstruct the output, or even reconstruct the image without its anomalies.
This idea is illustrated in Figure~\ref{fig:basic_autoencoder}.

We performed experiments comparing our proposed method to other state-of-the-art reconstruction-based anomaly detection methods \citep{Defard2021,Shi2021, Cohen2020, Wang2021} that achieved good performance on the MVTec-AD dataset \citep{Bergmann}, a general anomaly detection image dataset.
In experiments performed on a dataset containing PCB images under varied illumination conditions and camera angles, our method outperformed these state-of-the-art techniques, producing a more precise segmentation of the modifications and obtaining better scores on the measured metrics --- pixel-wise intersection over union (IoU), precision, recall, F-score, and detection and segmentation area under the receiver operating characteristic curve (AUROC).
Additionally, on the more general MVTec-AD dataset, our method performed similarly to the other methods, achieving better results for the addition or removal of objects.

The main contributions of this work are:

\begin{itemize}

\item { 
{We propose a loss function that combines the content loss concept and the mean squared error function for training a denoising convolutional autoencoder architecture for reconstruction-based anomaly detection.
The proposed model can be trained using only anomaly-free images, making it suitable for real-world applications where this kind of sample is much more common and easier to obtain than a representative set of samples containing anomalies.}}

\item { 
{We propose a comparison function that can be used to locate and segment regions that differ between a given input image and the reconstructed image produced by a convolutional autoencoder.
The comparison is based on higher level features instead of individual pixels, leading to the detection of structures and components instead of sparse noise.
}}

\item {  
{
We employ the proposed loss and comparison functions to design a robust method to detect modifications on PCBs that can be applied to images containing perspective distortion, noise, and lighting variations.
Thus, the method aims to work under the circumstances commonly found in practice, e.g.,  during the on-site inspection of gas pump PCBs \citep{Oliveira2017}, 
where mobile devices are used to capture images without relying on controlled lighting or positioning.
Nonetheless, it is important to highlight that the proposed method
may also be applied to other monitoring tasks that share similar characteristics, such as quality assurance in an industrial setting.}}

\item {We provide a labeled PCB image dataset for training and evaluating anomaly detection and segmentation methods. The dataset is publicly available\footnote {
\url{https://github.com/Diulhio/pcb_anomaly/tree/main/dataset}}and contains 1742 4096$\times$2816-pixel images from an unmodified gas pump PCB , as well as 55 images containing modifications, along with corresponding segmentation masks.}

\end{itemize}

The remainder of this paper is organized as follows. Section \ref {section.related} discusses related work on defect and anomaly detection on PCBs, as well as anomaly detection for industrial inspection in general.
Section \ref {section.method} details the proposed approach.
Section \ref {section.experiments} presents the experimental setup and the obtained results.
Finally, Section \ref {section.conclusion} draws some conclusions and indicates directions for future work.

%% file: 2_related_work.tex
\section{Related work}
\label {section.related}

Several algorithms have been proposed for image-based anomaly detection in PCBs.
For instance, deep learning techniques have been used to detect anomalies such as missing holes and defective circuits in unassembled boards \citep{Adibhatla2020,Shi2020}.
Although these approaches were successful, they rely on controlled capture conditions, and only work for limited types of anomalies, found in unassembled boards.
For assembled boards, a common strategy is using supervised training to produce a component detector \citep{Li2020,MallaiyanSathiaseelan2021}.
The layout of the detected components can be compared to a reference, providing a way of detecting anomalies, but this strategy demands considerable effort to obtain labeled training data (for example, \citep{Li2020} generates artificial samples from 3D models).
Moreover, this strategy is limited to detecting known components, possibly failing when the modification involves adding some unknown component.

Of particular relevance is the system proposed in \citep{Oliveira2017}, which addresses the same problem as we do.
We employed the same method used by that work to deal with variations in camera angle, and the same idea of partitioning the board to analyze each region independently.
Our main test dataset includes some of the images used by that work.
However, our anomaly detection strategy differs significantly --- they employ SIFT features and Support Vector Machines to classify each region as normal or anomalous, while we segment anomalies using a deep reconstruction network.
Moreover, that work uses supervised learning, with anomalies being artificially created by placing small patches extracted from other samples, while our model is semi-supervised, being trained only on normal samples.

Several semi-supervised methods have been recently proposed, which rely only on normal samples for anomaly detection for industrial visual inspection (not limited to PCBs).
The most successful methods are based on reconstructions or embedding similarity.
Reconstruction-based methods compute a compressed representation of the input image and attempt to reconstruct the original image based on it.
Our method falls into this category.
Models that can be employed for the reconstruction include autoencoders (AE) \citep{Bergmann, Shi2021}, variational autoencoders (VAE) \citep{Venkataramanan2020, Sato2019} and generative adversarial networks (GAN) \citep{Akcay2018}.
The main advantage of these approaches is that it is easy for a human to understand and interpret their results.
However, if a method still reconstructs an anomaly \citep{Perera2019}, it may remain undetected, as there is no noticeable difference between the input and the reconstruction.

Embedding similarity methods \citep{Cohen2020, Defard2021, Wang2021} use deep convolutional networks pre-trained on large generic datasets (e.g.~ImageNet) as feature extractors.
The distribution of the features extracted from anomaly-free samples is then modeled as a probability density function \citep{Cohen2020}.
Given a distance metric, the feature vectors from images with anomalies tend to be more distant to the center of the distribution (e.g.~the mean vector), compared to normal samples.
These methods are applicable to new problem domains without requiring additional training of the basic feature extractor, but their results are hard to interpret.
Moreover, computation of the density function can have high memory requirements and be complicated when the dataset has high variability.

A popular benchmark for visual anomaly inspection is the MVTec-AD \citep{Bergmann, Bergmann2021} dataset, which contains 5354 images, with 70 types of anomalies for 15 kinds of objects.
Most of the anomaly detection methods cited above were evaluated on this dataset, so our method will also be tested on it.

%% file: 4_method.tex
\section{Proposed method}
\label {section.method}

Many existing approaches for anomaly detection produce a binary classification that refers to the entire sample, telling whether it contains a modification or not.
However, this may be insufficient in a real-world scenario, since the specific structures or components which characterize the modification are not identified.
Methods that produce bounding boxes or segment anomalies may be more suitable for PCB inspection.
Thus, the approach we propose in this work performs anomaly segmentation.
It employs a deep convolutional neural network for image reconstruction, trained in a semi-supervised manner, only on normal samples, i.e.~images without modifications/defects/anomalies.


\subsection {Image registration and partitioning}
\label {section.registration_and_partitioning}

\begin{figure}[]
    \centering
    \begin{subfigure}{.48\linewidth}
        \centering
        \includegraphics[width=\linewidth]{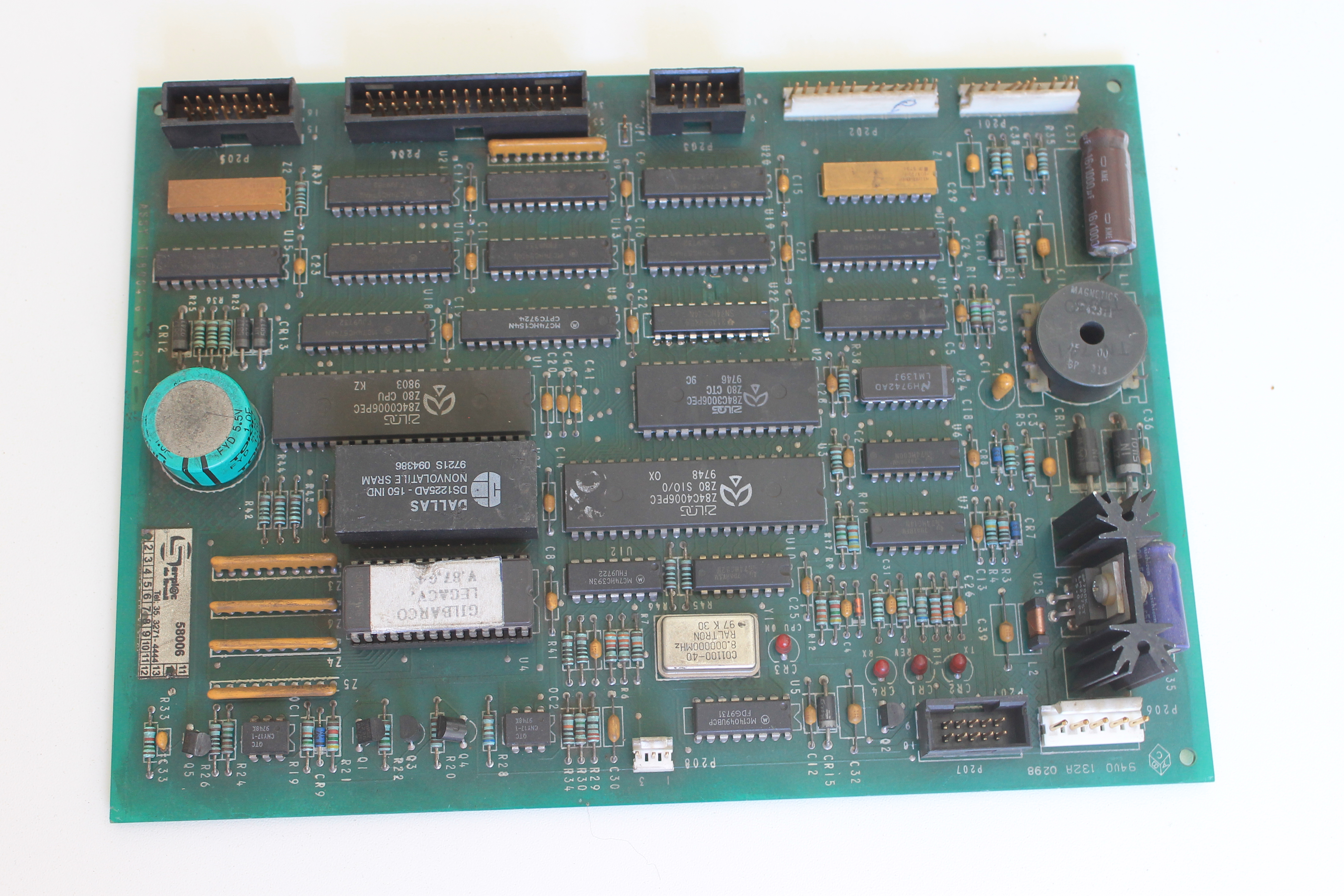}
        \caption{}
        \label{fig:registration2}
    \end{subfigure}
    \begin{subfigure}{.48\linewidth}
        \centering
        \includegraphics[width=\linewidth]{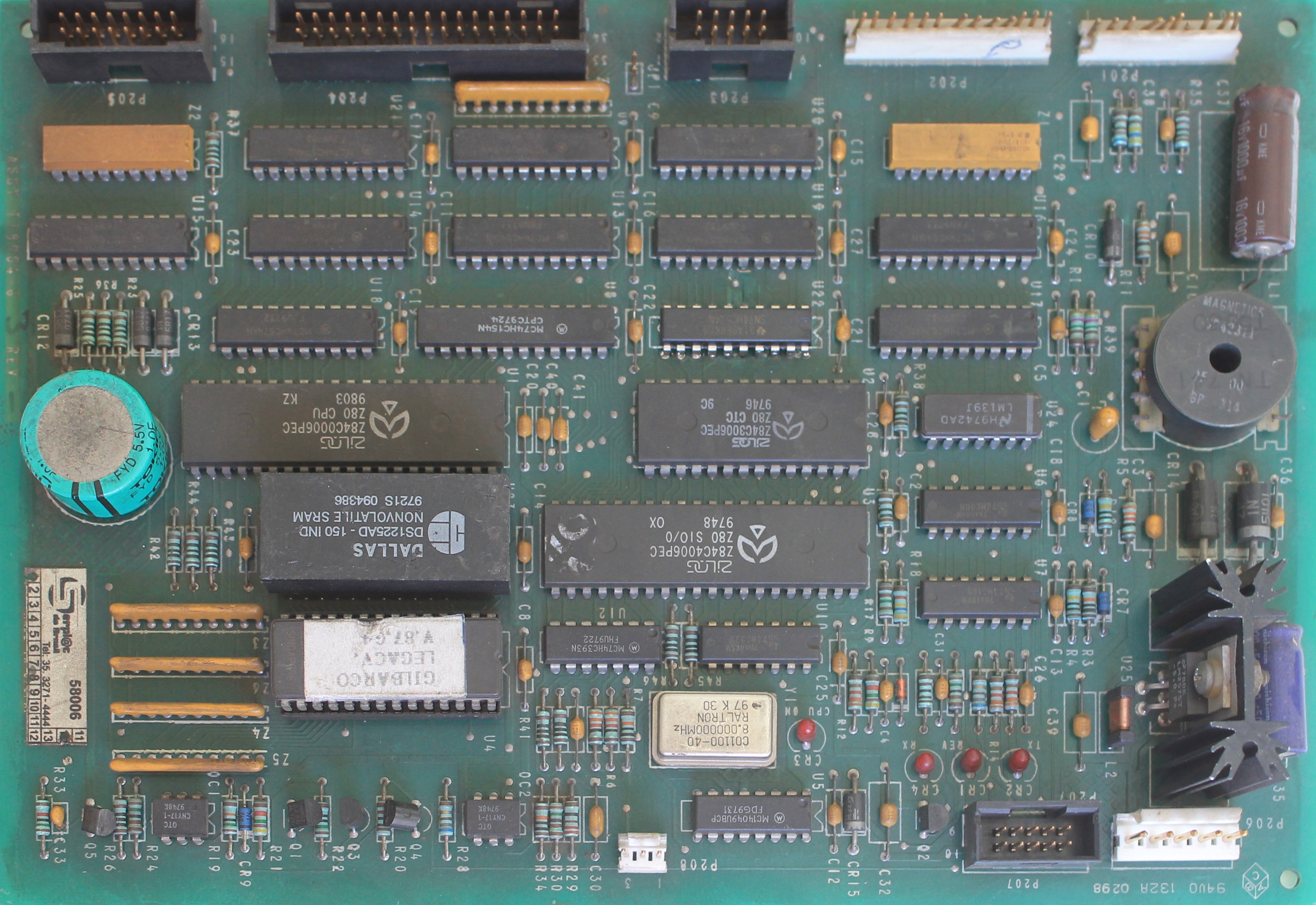}
        \caption{}
        \label{fig:registration1}
    \end{subfigure}
    \caption{In (a) the original image, and (b) the image after registration using SIFT and RANSAC.}
    \label{fig:registration}
\end{figure}

\begin{figure}[!htbp]
    \centering
    \begin{subfigure}{.31\linewidth}
        \centering
        \includegraphics[width=\linewidth]{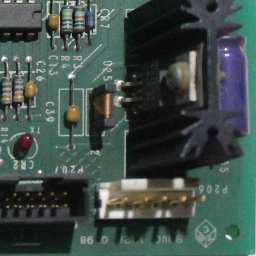}
    \end{subfigure}
    \begin{subfigure}{.31\linewidth}
        \centering
        \includegraphics[width=\linewidth]{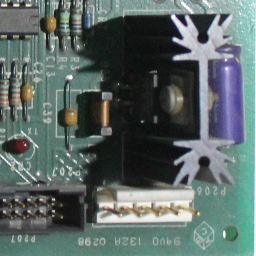}
    \end{subfigure}
    \begin{subfigure}{.31\linewidth}
        \centering
        \includegraphics[width=\linewidth]{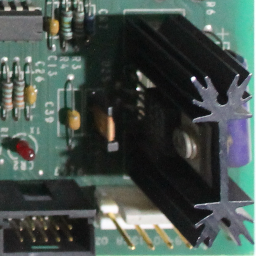}
    \end{subfigure}
    \begin{subfigure}{.31\linewidth}
        \centering
        \includegraphics[width=\linewidth]{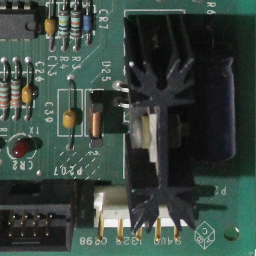}
    \end{subfigure}
    \begin{subfigure}{.31\linewidth}
        \centering
        \includegraphics[width=\linewidth]{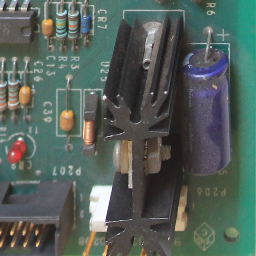}
    \end{subfigure}
    \begin{subfigure}{.31\linewidth}
        \centering
        \includegraphics[width=\linewidth]{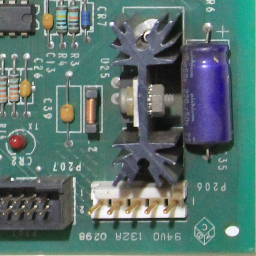}
    \end{subfigure}
    \caption{Taller components can have considerable aspect variation even after the image registration procedure.}
    \label{fig:comp_variation}
\end{figure}

Similarly to the work from \citep{Oliveira2017}, our approach assumes the PCB is shown from an overhead view.
However, different from several other studies on visual inspection \citep{Bergmann2021, Wang2021, Adibhatla2020, Li2020}, where positioning is strict to avoid variations, we suppose the input image may be the product of an image registration step.
In other words, the PCB may be photographed from an angled view, being aligned to a reference image after capture (see Figure \ref {fig:registration}).
{We employed a widely used and mature algorithm for image registration based on SIFT features and the RANSAC algorithm \citep{Lowe2004}, but note that any algorithm with good performance could be used.
More relevant for our discussion are the implications of relying on an image registration step:}
in the resulting image, the components on the PCB may have some degree of perspective distortion and variations in position, since image registration can be slightly imprecise, and the algorithm treats only planar distortions, without taking into account the 3D aspect of the components, as shown in Figure~\ref {fig:comp_variation}.
Moreover, our approach does not demand controlled lighting, so there can be reflections, shadows and other variations, which can be hard to distinguish from actual modifications or anomalies.
These assumptions make our approach suitable for real-world applications where the inspection may occur in an open and uncontrolled environment.

\begin{figure}[!htbp]
    \centering
    \begin{subfigure}{.24\linewidth}
        \centering
        \includegraphics[width=\linewidth]{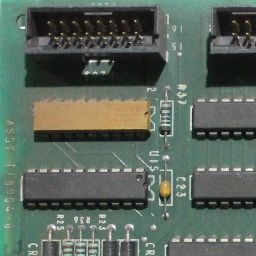}
    \end{subfigure}
    \begin{subfigure}{.24\linewidth}
        \centering
        \includegraphics[width=\linewidth]{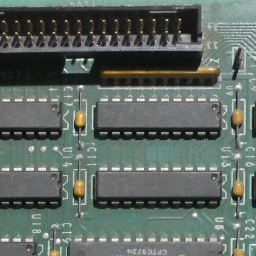}
    \end{subfigure}
    \begin{subfigure}{.24\linewidth}
        \centering
        \includegraphics[width=\linewidth]{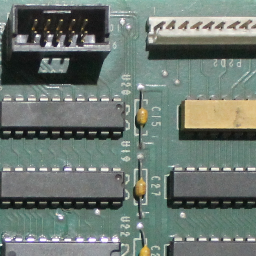}
    \end{subfigure}
    \begin{subfigure}{.24\linewidth}
        \centering
        \includegraphics[width=\linewidth]{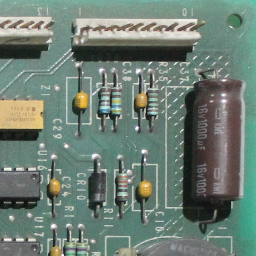}
    \end{subfigure}
    
    \begin{subfigure}{.24\linewidth}
        \centering
        \includegraphics[width=\linewidth]{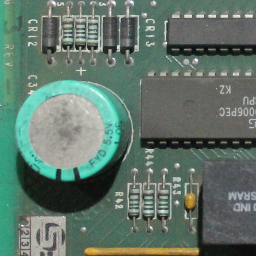}
    \end{subfigure}
    \begin{subfigure}{.24\linewidth}
        \centering
        \includegraphics[width=\linewidth]{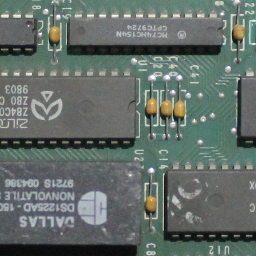}
    \end{subfigure}
    \begin{subfigure}{.24\linewidth}
        \centering
        \includegraphics[width=\linewidth]{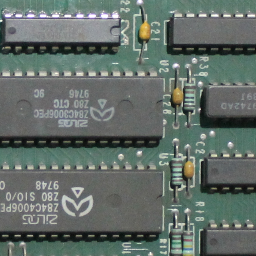}
    \end{subfigure}
    \begin{subfigure}{.24\linewidth}
        \centering
        \includegraphics[width=\linewidth]{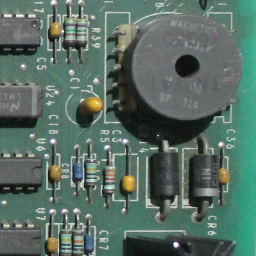}
    \end{subfigure}
    
    \begin{subfigure}{.24\linewidth}
        \centering
        \includegraphics[width=\linewidth]{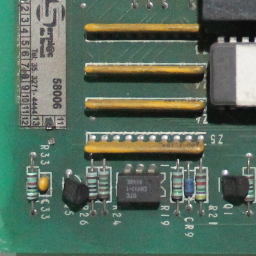}
    \end{subfigure}
    \begin{subfigure}{.24\linewidth}
        \centering
        \includegraphics[width=\linewidth]{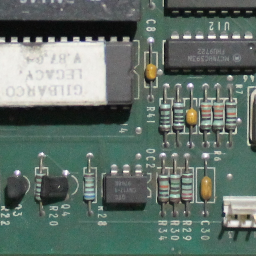}
    \end{subfigure}
    \begin{subfigure}{.24\linewidth}
        \centering
        \includegraphics[width=\linewidth]{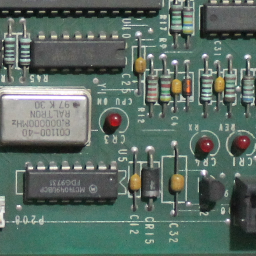}
    \end{subfigure}
    \begin{subfigure}{.24\linewidth}
        \centering
        \includegraphics[width=\linewidth]{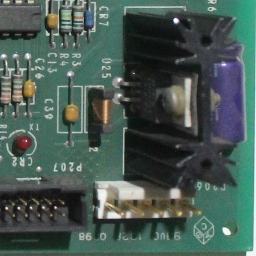}
    \end{subfigure}

    \caption {A 4096$\times$2816 image split into 1024$\times$1024 patches. Some regions are present in more than one patch --- the regions overlap because 2816 is not divisible by 1024.}
    \label{fig:img_grid}
\end{figure}

Anomaly detection methods frequently work on fixed-size inputs, reducing the captured image to a smaller size, which also reduces computation and memory requirements.
However, for PCB inspection in the proposed dataset, resizing the entire image to a manageable size can result in certain components and modifications becoming too small.
To avoid this, we partition the input image into 1024$\times$1024-pixel patches (to avoid having an overly large number of patches per image), which are then resized to 256$\times$256 pixels (to reduce computational costs) and processed independently.
Figure~\ref{fig:img_grid} illustrates this procedure.

\subsection{Convolutional autoencoder architecture}

After the original image is partitioned, each 256$\times$256 patch is given as an input to a convolutional autoencoder (CAE) \citep{Courville2016a}.
Using a series of convolutional layers, CAEs encode the high-dimensional input image to a compressed low-dimensional vector called ``latent space'' and expand (decode) this vector to the original dimensionality.
The encoder function $z = g_{\phi}(x)$ receives the input and maps it to the latent space $z$.
The decoder function $x' = f_{\theta}(z)$ computes the reconstruction from the latent space.
Thus, the entire network is expressed as $f_{\theta}(g_{\phi}(x)) = x'$, and in a perfect CAE $x = x'$.

In our approach, one CAE is trained for each patch region (i.e.~for the board shown in Figure \ref {fig:img_grid}, we have 12 CAEs).
These networks are trained using only anomaly-free samples, ideally becoming able to reconstruct only this type of image --- when receiving images showing anomalies, the CAE will produce visible artifacts or reconstruct them without the anomalies, as illustrated in Fig.~\ref {fig:basic_autoencoder}.

The CAE architecture we use in our approach is shown in Table~\ref{table:architecture}.
The network was built using convolutional layers in the encoder and transposed convolutional layers in the decoder, with 5$\times$5 kernels in both cases.
Each convolutional layer is followed by Batch Normalization (BN) and a Leaky ReLU activation, with a slope of 0.2.
The last layer of the encoder and the first layer of the decoder are fully connected layers of 1024 nodes, followed by BN and Leaky ReLU.
The latent space is the output of a fully connected layer with 500 values.

\begin{table}[]
\begin{tabular}{l}
\hline \hline
Input: $x_{256\times256\times3}$	\\ \hline \hline
Conv(filters=32); BN; LeakyReLU  		\\ \hline
Conv(filters=64); BN; LeakyReLU  		\\ \hline
Conv(filters=128); BN; LeakyReLU 		\\ \hline
Conv(filters=128); BN; LeakyReLU 		\\ \hline
Conv(filters=256); BN; LeakyReLU 		\\ \hline
Conv(filters=256); BN; LeakyReLU 		\\ \hline
Conv(filters=256); BN; LeakyReLU 		\\ \hline
Fully connected (1024); BN; Leaky ReLU  \\ \hline
Fully connected (500); Leaky ReLU       \\ \hline
Fully connected (1024); BN; Leaky ReLU  \\ \hline
TranspConv(filters=256); BN; LeakyReLU 		\\ \hline
TranspConv(filters=256); BN; LeakyReLU 		\\ \hline
TranspConv(filters=128); BN; LeakyReLU 		\\ \hline
TranspConv(filters=128); BN; LeakyReLU 		\\ \hline
TranspConv(filters=64); BN; LeakyReLU  		\\ \hline
TranspConv(filters=32); BN; LeakyReLU  		\\ \hline
TranspConv(filters=3); Sigmoid          		\\ \hline
\end{tabular}
\centering
\caption{The architecture of our convolutional autoencoder. All convolutional and transposed convolutional layers use 5$\times$5 kernels and stride 2.}
\label{table:architecture}
\end{table}

{During training, each input image is corrupted by randomly masking out rectangular regions --- denoising autoencoders use this data corruption strategy to prevent the network from simply memorizing the training data. The effect is similar to dropout, but in input space --- generating images with simulated occlusions forces the model to take more of the image context into consideration when extracting features, improving network generalization \citep{DeVries2017}.
Note that the loss is still computed by comparing the produced output to the original, non-corrupted input.}


\subsection{Content loss function for training}

The loss functions most commonly used for training autoencoders are pixel-wise functions, such as the Mean Square Error (MSE).
However, these functions assume the pixels are not correlated, which is often not true --- in general, images have structures formed by the relations between pixel neighborhoods.
Pixel-wise functions also frequently result in blurred outputs when used for reconstruction.
For these reasons, we used the content loss function when training the autoencoder.

Content loss, introduced by \citep{Gatys2016}, identifies differences between two images (in our case, the input and the reconstruction) based on high-level features.
It was used for applications such as style transfer \citep{Gatys2016, Johnson2016}, super-resolution \citep{Johnson2016, Ledig2017} and image restoration \citep{Zhao2017}.
Features are extracted from an image classification network (VGG19 \citep{Liu2015}, in our work) pre-trained on general-purpose datasets (Imagenet \citep{Deng2009}, in our work).
This function encourages the network to reconstruct images with feature representations similar to those of input, rather than considering just differences between pixels.

Let $\phi_{j}(x)$ be the activation of the $j$th layer of a pre-trained network $\phi$ when image $x$ is processed.
Since $j$ is a convolutional layer, $\phi_{j}(x)$ will present an output of shape $ C_{j}\times H_{j}\times W_{j}$, where $C_{j}$ is the number of filter outputs, and $H_{j}\times W_{j}$ is the size of each filter output at layer $j$.
The content loss is the squared and normalized distance of the feature representations of the reconstruction $\hat{y}$ and the reference $y$, as expressed in Eq.~\ref{eq:perploss}.

\begin{equation}
\label{eq:perploss}
l^{\phi,j}_{feat}(\hat{y}, y) = \frac{1}{C_{j}H_{j}W_{j}} \left \| \phi_{j}(\hat{y}) - \phi_{j}(y) \right \|^{2}_{2}
\end{equation}

This function tries to find an image $\hat{y}$ that minimizes the reconstruction loss using the initial layers of the pre-trained network $\phi$.
A CAE trained with this function tends to produce images similar to target $y$ in image content and overall spatial structure \citep{Johnson2016}.
In this work, we sum the differences in the 5th, 8th, 13th and 15th layers from VGG19, based on empirical experiments.

\begin{figure}[!htbp]
    \includegraphics[width=\linewidth]{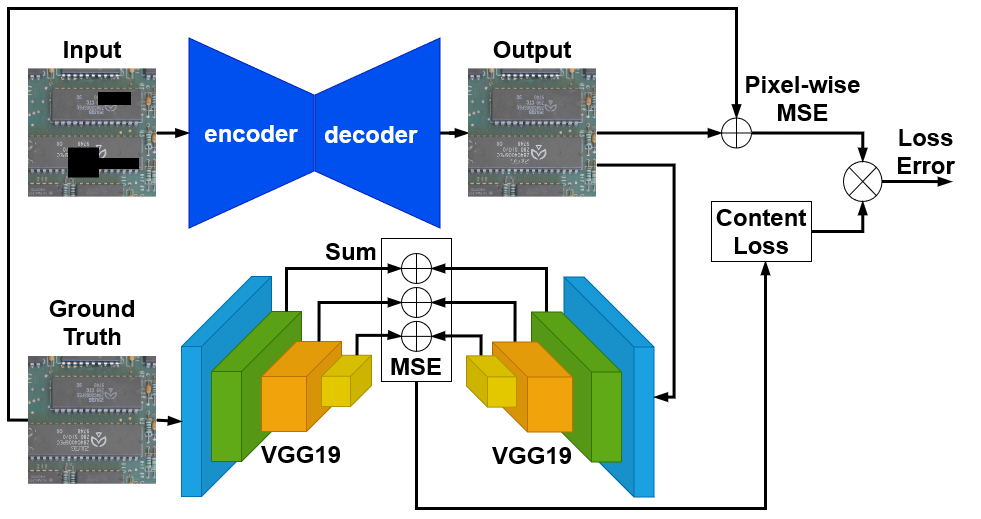}
    \caption{The loss calculation flow during training.}
    \label{fig:loss_diagram}
\end{figure}

The content loss function controls the reconstruction of larger structures in the image but fails to reconstruct details and textures.
For this reason, we combine content loss with the MSE, as expressed by Eq.~\ref{eq:loss_all}, where $\lambda_{1}$ and $\lambda_{2}$ are the weights of each loss function.
We empirically defined the parameters $\lambda_{1}=0.01$ and $\lambda_{2}=1$. Figure~\ref{fig:loss_diagram} illustrates the entire loss calculation.

\begin{equation}
\label{eq:loss_all}
\mathcal{L}_{rec} = \lambda_{1} \mathcal{L}_{MSE} + \lambda_{2} \mathcal{L}_{feat} 
\end{equation}

\subsection{Anomaly segmentation}
\label{section.segmentation}

\begin{figure*}[]
    \centering
    \begin{subfigure}{.15\linewidth}
        \begin{subfigure}{\linewidth}
            \centering
            \includegraphics[width=\linewidth]{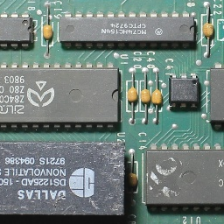}
        \end{subfigure}
        \begin{subfigure}{\linewidth}
            \centering
            \includegraphics[width=\linewidth]{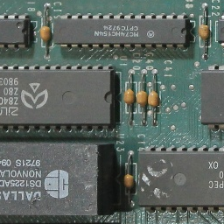}
        \end{subfigure}
        \vspace{4mm}
        \caption{}
    \end{subfigure}
    \begin{subfigure}{.15\linewidth}
        \centering
        \begin{subfigure}{\linewidth}
            \centering
            \includegraphics[width=\linewidth]{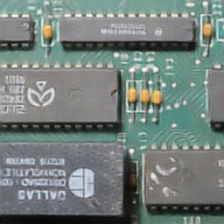}
        \end{subfigure}
        \begin{subfigure}{\linewidth}
            \centering
            \includegraphics[width=\linewidth]{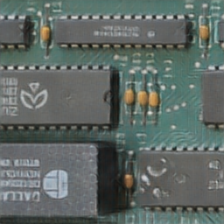}
        \end{subfigure}
        \vspace{4mm}
        \caption{}
    \end{subfigure}
    \begin{subfigure}{.15\linewidth}
        \begin{subfigure}{\linewidth}
            \centering
            \includegraphics[width=\linewidth]{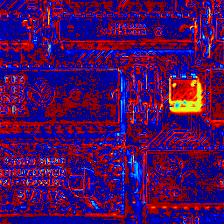}
        \end{subfigure}
        \begin{subfigure}{\linewidth}
            \centering
            \includegraphics[width=\linewidth]{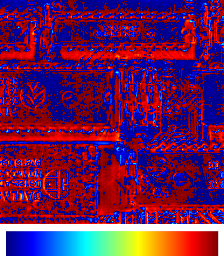}
        \end{subfigure}
        \caption{}
    \end{subfigure}
    \begin{subfigure}{.15\linewidth}
        \begin{subfigure}{\linewidth}
            \centering
            \includegraphics[width=\linewidth]{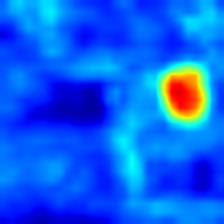}
        \end{subfigure}
        \begin{subfigure}{\linewidth}
            \centering
            \includegraphics[width=\linewidth]{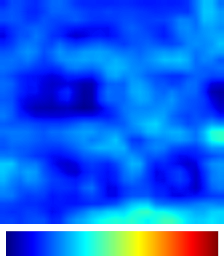}
        \end{subfigure}
        \caption{}
    \end{subfigure}
    
    \caption{
    {(a) Input image showing a PCB with (top) and without (bottom) modifications, (b) reconstruction produced by the autoencoder network, (c) the pixel-wise absolute difference between the input and its reconstruction, and (d) the proposed anomaly segmentation method using perceptual difference. The absolute difference shows high values spread around many areas, even where there are no modifications. With the proposed method, the region containing the anomaly has markedly higher values than regions without anomalies.}}
    \label{fig:abs_diff}
\end{figure*}

After training, the network can be used to segment anomalies by comparing the reconstructed image to the input.
If the CAE was ``perfect'', a simple pixel-wise absolute difference would be enough to segment the anomalies.
However, images in a real situation have perspective distortion, noise, and lighting variations that may make the reconstruction hard.
These variations may cause small differences along edges, or in regions containing shadows or reflections.
In these cases, pixel-wise metrics may result in many false positives.
Figure~\ref{fig:abs_diff}(c) shows an example of the absolute difference between the original images of a PCB (with and without modifications) and their reconstructions.
The pixel-wise absolute difference has high values at several positions, even in places where the differences are very hard to notice.

To address these challenges, we propose a comparison function based on the content loss concept, i.e.~instead of isolated pixels, we focus on structures and higher level features.
{Tiny modifications that manifest in isolated pixels may pass undetected, but the overall robustness is increased, since actual modifications to PCBs appear as clusters of pixels, as long as the board is photographed with a good enough resolution.}
Once again, we used the VGG19 network trained on the ImageNet dataset to extract high-level features from the input $y$ and the reconstruction $\hat{y}$.
The features are compared by summing the absolute differences between the activations of layer $\phi_{j}$, as expressed in Eq.~\ref{eq:ano_map}.
\begin{equation}
\label{eq:ano_map}
\mathcal{A}(\hat{y}, y) = \sum_{i}^{C_{j}} \left | \phi_{j, i}(\hat{y}) - \phi_{j, i}(y) \right |
\end{equation}

where $C_{j}$ is the number of filter outputs in layer $j$. 
$\mathcal{A}$ is a matrix that represents the anomaly map, and has the same size ($H_{j}\times W_{j}$) as the outputs from layer $\phi_{j}$.
In initial tests performed on a small dataset, the 12nd layer from VGG19 showed the best results, with 512 outupts of size 28$\times$28.

To get the final segmentation, the anomaly map is resized using bilinear interpolation to the same size as the input, normalized, and binarized with a threshold $T$.
Normalization is based on the min-max range from the entire test set, which must contain images showing modifications, so that we have a measure of the magnitudes of the values produced by these anomalies.
The $T$ parameter gives a measure of how rigorous the detection is, and will be varied during the experiments.
Figure~\ref{fig:abs_diff}(d) shows an example of the proposed segmentation method.
Note how differences in regions without anomalies are much less noticeable than when using the pixel-wise absolute difference. 
On the other hand, the region containing the anomaly has much higher values in the anomaly map than other regions.

%% file: 5_experiments.tex
\section{Experiments and results}
\label {section.experiments}

In this section, we present the experiments performed to test the proposed approach for anomaly detection and compare it with other semi-supervised state-of-art methods, on our MPI-PCB dataset and the MVTec-AD dataset.
Everything was implemented in the Python language, using the TensorFlow\footnote {www.tensorflow.org} and OpenCV\footnote {opencv.org} libraries.
Experiments were performed on the Google Colab\footnote {colab.research.google.com} platform.
The source code is publicly available at \url{https://github.com/Diulhio/pcb_anomaly/}.


\subsection{MPI-PCB Dataset}

The main dataset used in this work is the \textbf{M}ulti-\textbf{P}erspective and \textbf{I}llumination PCB (MPI-PCB) dataset, which we built based on many of the same images originally collected for the work in \citep{Oliveira2017}.
The dataset contains 1742 4096$\times$2816-pixel images showing an unmodified PCB from a gas pump.
The images were captured using a Canon EOS 1100D camera with 18-55mm lenses.
The set also contains 55 images showing the board with modifications manually added by the authors, which are meant to be representative of situations encountered in actual frauds.
As usual in semi-supervised training, these samples must not be used in the training step, only for testing.
One of the contributions of our paper is making this dataset available, including labeled semantic segmentation masks. 

Images were captured from a generally overhead view, but without strict demands on position or illumination, as expected in a real-world situation.
To reduce variations that may occur in the image registration step and focus on the anomaly detection problem, the dataset contains the images after the registration procedure described in Section \ref {section.registration_and_partitioning}.


\subsection{Baseline Methods}

To the best of the authors' knowledge, there are no previous work addressing specifically image-based anomaly segmentation in assembled PCBs --- as previously discussed in Section \ref {section.related}, existing approaches focus on unassembled PCBs or use supervised training to determine if anomalies are present in a given region, without per-pixel segmentation. 
That makes it hard to directly compare this kind of approach with the proposed method. 
For that reason, our comparisons are focused on other general anomaly segmentation methods, which achieved promising results on the popular MVTec-AD dataset.
Our work can be compared to these methods more directly, since they have similar semi-supervised training procedures and produce segmentation masks as outputs.
We chose baseline methods that provide the source code and can run in the infrastructure used for our work.
We also selected at least one reconstruction-based method and one embedding similarity method.

Up to the time our experiments were performed, the PaDiM approach \citep{Defard2021} had the state-of-art results for anomaly segmentation on the MVTec-AD dataset.
It is an embedding similarity method that obtained the best results when using the Wide ResNet-50-2 network to extract features, but due to the very high memory requirements, we used the smaller ResNet18 as a feature extractor in our comparison.
Other embedding similarity methods we used as baselines were SPTM \citep{Wang2021} and SPADE \citep{Cohen2020}.
For the latter, we reduced the input resolution from the default 224 $\times$ 224 to 192 $\times$ 192, also due to the high memory requirements.
As a reconstruction-based baseline, we took the DFR method \citep{Shi2021}, which uses regional features extracted from a pre-trained VGG19 as inputs for CAEs.


\subsection{Evaluation metrics}

We considered per-pixel metrics to evaluate the segmentation performance of the techniques: the intersection over the union (IoU) and the area under the receiver operating characteristic curve (ROC-AUC), as well as the usual precision, recall and F-score in the best case.
We also evaluated ROC-AUC for anomaly detection: while segmentation considers per-pixel classification, detection expresses if an anomaly exists or not in the image.
To avoid detecting noise, we consider an anomaly exists in an image if it contains at least 10 anomalous pixels.
The metrics are computed over the (per pixel or per image) count of true positive (TP), true negative (TN), false positive (FP), and false negative (FN) classifications.

Precision indicates the proportion of detected pixels that were correct, i.e.~values close to 1 indicate there were few false detections; while recall indicates the proportion of expected pixels that were detected, i.e.~values close to 1 indicate that most of the anomalies were detected.
More formally, precision (eq.~\ref{eq:precision}) expresses the ratio of correctly predicted positive samples to the total predicted positive samples; and
recall (eq.~\ref{eq:tpr}), also known as true positive rate (TPR), expresses the ratio of correctly predicted positive samples to all the samples in the positive class.
The $F$-score (eq.~\ref{eq:f1}) is the harmonic mean of precision and recall.

\begin{equation}
\label{eq:precision}
Precision = \frac{TP}{TP+FP}
\end{equation}

\begin{equation}
\label{eq:tpr}
Recall/TPR = \frac{TP}{TP+FN}
\end{equation}

\begin{equation}
\label{eq:f1}
\text{\textit{F-score}} = \frac{2*(Recall * Precision)}{(Recall + Precision)}
\end{equation}

ROC-AUC is a widely used metric for evaluating anomaly segmentation methods, and is usually reported for approaches tested on the MVTec-AD dataset \citep{Bergmann, Cohen2020, Bergmann2021, Defard2021, Wang2021, Shi2021}.
It shows how well a technique balances true and false positive rates (i.e.~its ability to cover the expected detections while avoiding false detections) as a certain threshold parameter varies.
ROC-AUC is the normalized area under the ROC curve, which is the curve obtained by plotting the true versus the false positive rates (TPR and FPR, respectively) at different classification thresholds.
TPR and FPR are computed by Equations~\ref{eq:tpr} and \ref{eq:fpr}.

\begin{equation}
\label{eq:fpr}
FPR = \frac{FP}{FP+TN}
\end{equation}

IoU, also referred to as the Jaccard index, is also reported for several semantic segmentation tasks and challenges such as COCO\footnote {Common Objects in Context – http://cocodataset.org.}.
For anomaly segmentation, the IoU expresses how similar two shapes are, quantifying the overlap between the ground truth mask and the binarized anomaly map, as given by Equation~\ref{eq:iou}.

\begin{equation}
\label{eq:iou}
IoU = \frac{TP}{(TP + FN + FP)}
\end{equation}

We report the best IoU score obtained by each method when varying the classification threshold. Compared to the ROC-AUC, the IoU is more sensitive to variations in the shape of the segmented regions.


\subsection {Training details}

We took the 1742 images from the MPI-PCB dataset showing the unmodified board, and randomly split them as follows: 1518 images for training, 169 for validation, and 55 for testing (the same amount we have with the modified board, to a total of 110 test images). As for the MVTec-AD dataset, the split is: 3266 for training, 363 for validation, and 1725 for test \citep{Bergmann}.

Due to the large variety of perspective distortions, as well as the limited number of training samples, we used data augmentation on the training sets from both datasets.
For the MPI-PCB dataset, we apply a random position offset between 0 and 80 pixels when extracting patches, simulating variations that may occur in the image registration step.
As for the MVTec-AD dataset, we apply random variations on rotation, shear, saturation, contrast, brightness and scale.

The proposed architecture was trained with a batch size of 128 for 1000 epochs.
As optimizer, we used Adam with cosine learning rate decay and a warm-up phase.
The learning rate starts at $1 \times 10^{-5}$, after 3 epochs ramps up to 0.0072 and decays to $1 \times 10^{-5}$ using a cosine function.


\subsection{Results on the MPI-PCB dataset}

We evaluated the performance of our method and of the baseline methods on the test set from the MPI-PCB dataset, considering six board regions.
All these regions contain inserted modifications, like integrated circuits and jumper wires.
The tested methods depend on at least part of test samples from each region containing modifications, to define the range for normalization.
A total of 110 samples were tested, 55 with and 55 without modifications in the observed region.

\begin{table*}[!ht]
\small
\centering
\caption{Results of the proposed method and the baseline methods for the MPI-PCB dataset. We show the results for image regions containing at least one anomaly in the test set. The grid numbers indicate the column/row of each region in the partitioned image (see Fig.~\ref{fig:img_grid}). Higher values indicate better performance, where segmentation precision, recall and F-score are shown at the best IoU threshold.}
\label{table:results_mpi}
\begin{tabular}{llrrrrrrr}
\hline
\multicolumn{1}{c}{Metric}  & \multicolumn{1}{c}{Method} & \multicolumn{7}{c}{Region}                                                                                                                                                                                             \\ \cline{3-9} 
                            &                            & \multicolumn{1}{c}{grid2\_2} & \multicolumn{1}{c}{grid2\_3} & \multicolumn{1}{c}{grid3\_1} & \multicolumn{1}{c}{grid3\_2} & \multicolumn{1}{c}{grid4\_1} & \multicolumn{1}{c}{grid4\_3} & \multicolumn{1}{c}{Average} \\ \hline
\multirow{5}{*}{IoU}        & Ours                       &  \textbf{0.755}                        &  \textbf{0.664}                        &  \textbf{0.608}                        &  \textbf{0.525}                        &  \textbf{0.778}                        &  \textbf{0.732}                        &  \textbf{0.677}                       \\
                            & PaDiM                      & 0.603                        & 0.624                        & 0.489                        & 0.145                        & 0.656                        & 0.524                        & 0.507                       \\
                            & SPADE                      & 0.319                        & 0.272                        & 0.419                        & 0.353                        & 0.457                        & 0.474                        & 0.382                       \\
                            & DFR                        & 0.297                        & 0.098                        & 0.117                        & 0.386                        & 0.196                        & 0.190                        & 0.214                       \\
                            & SPTM                       & 0.505                        & 0.428                        & 0.447                        & 0.314                        & 0.240                        & 0.502                        & 0.406                       \\ \hline
\multirow{5}{*}{\shortstack[l]{Segmentation \\ Precision}}	& Ours                       & \textbf{0.858}               & \textbf{0.752}              & \textbf{0.767}              & \textbf{0.643}              & \textbf{0.849}              & \textbf{0.840}                       & \textbf{0.785}                       \\
							& PaDiM                       & 0.732                       & 0.742                       & 0.594                       & 0.240                       & 0.765                       & 0.687                       & 0.627                       \\
							& SPADE                       & 0.364                       & 0.301                       & 0.621                       & 0.417                       & 0.526                       & 0.533                       & 0.460                       \\
							& DFR                       & 0.246                       & 0.078                       & 0.117                       & 0.419                       & 0.141                       & 0.221                       & 0.204                       \\
							& SPTM                       & 0.601                       & 0.577                       & 0.457                       & 0.410                       & 0.300                       & 0.627                       & 0.495                       \\ \hline

\multirow{5}{*}{\shortstack[l]{Segmentation \\ Recall}}     & Ours                       & \textbf{0.876}               & \textbf{0.858}              & 0.758                       & \textbf{0.747}              & \textbf{0.915}              & \textbf{0.856}              & \textbf{0.835}                       \\
							& PaDiM                       & 0.856                       & 0.851                       & \textbf{0.826}              & 0.413                       & 0.896                       & 0.744                       & 0.764                       \\
							& SPADE                       & 0.754                       & 0.754                       & 0.572                       & 0.715                       & 0.793                       & 0.833                       & 0.737                       \\
							& DFR                       & 0.687                       & 0.491                       & 0.310                       & 0.691                       & 0.347                       & 0.407                       & 0.489                       \\
							& SPTM                       & 0.760                       & 0.853                       & 0.668                       & 0.643                       & 0.395                       & 0.760                       & 0.680                       \\ \hline

\multirow{5}{*}{\shortstack[l]{Segmentation \\ F-Score}}	& Ours                       & \textbf{0.863}               & \textbf{0.805}               & \textbf{0.769}             & \textbf{0.688}              & \textbf{0.889}              & \textbf{0.851}              & \textbf{0.811}              \\
							& PaDiM                       & 0.785                       & 0.791                       & 0.691                       & 0.307                       & 0.829                       & 0.714                       & 0.686                       \\
							& SPADE                       & 0.489                       & 0.436                       & 0.597                       & 0.521                       & 0.632                       & 0.645                       & 0.553                       \\
							& DFR                       & 0.357                       & 0.126                       & 0.163                       & 0.511                       & 0.209                       & 0.283                       & 0.275                       \\
							& SPTM                       & 0.676                       & 0.687                       & 0.533                       & 0.492                       & 0.347                       & 0.680                       & 0.569                       \\ \hline
\end{tabular}
\end{table*}

\begin{figure*}[!htbp]
    \includegraphics[width=\linewidth]{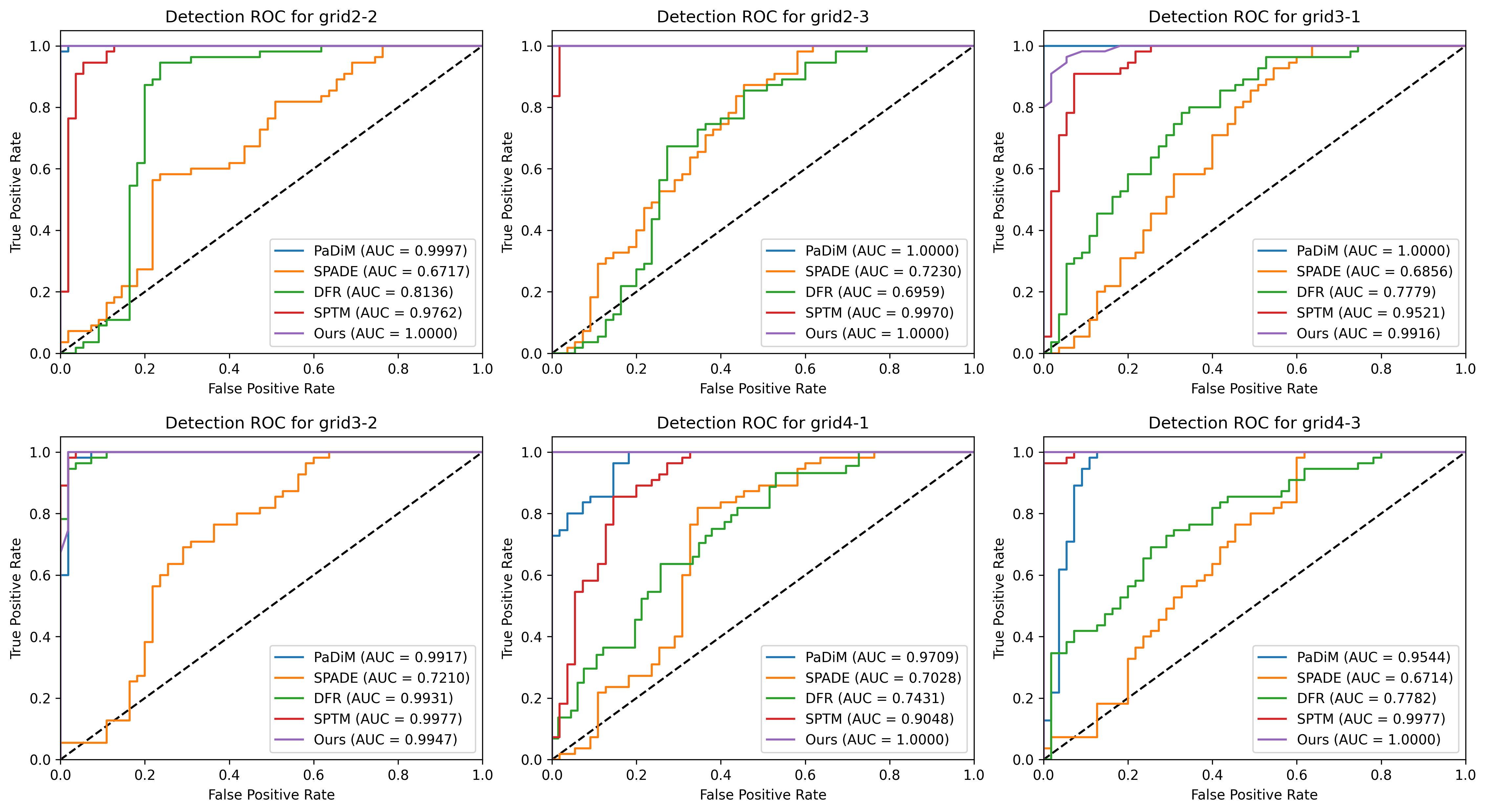}
    \caption{Detection ROC curves and AUC for each tested region.}
    \label{fig:mpi_det_curves}
\end{figure*}

\begin{figure*}[!htbp]
    \includegraphics[width=\linewidth]{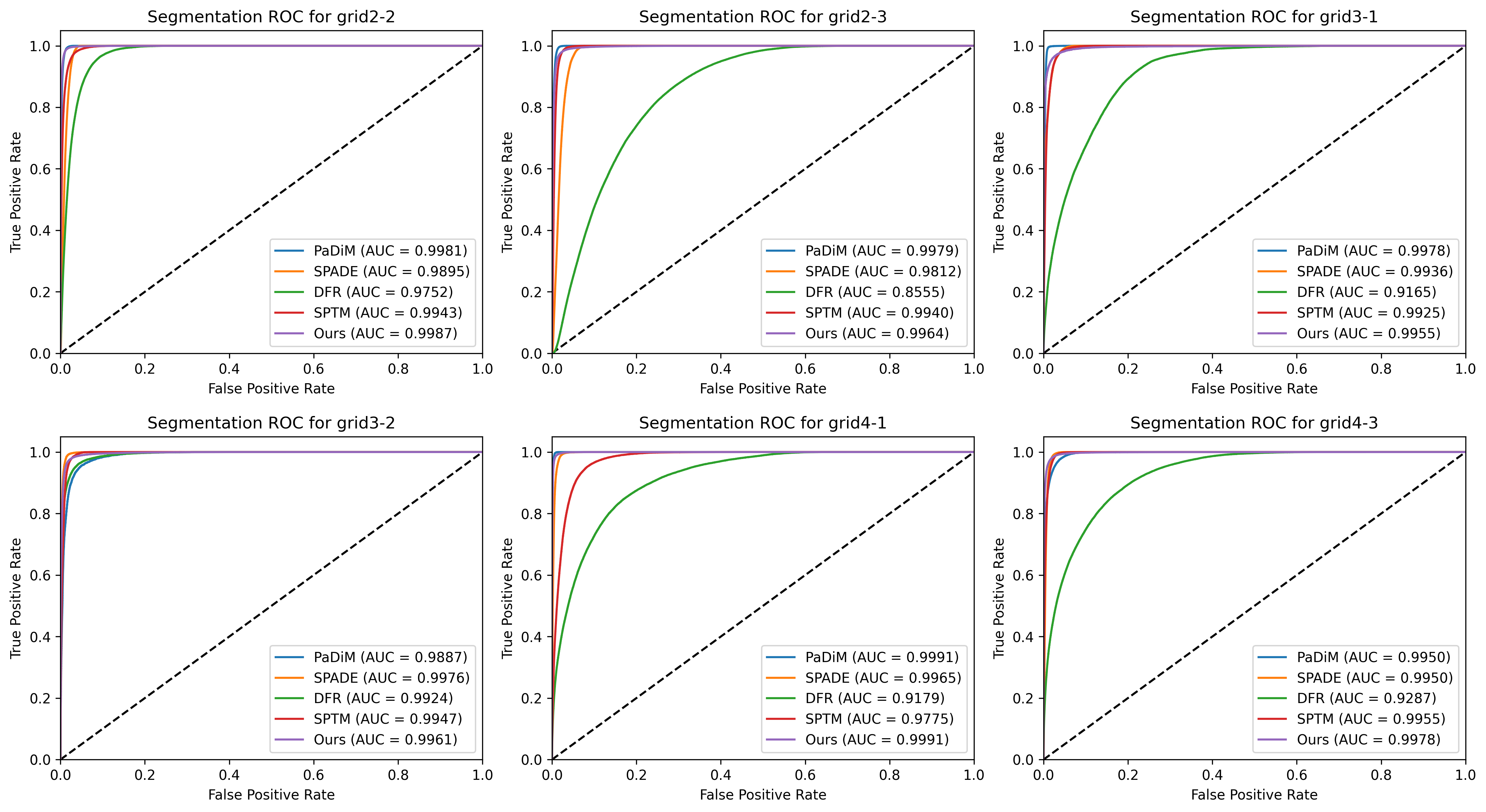}
    \caption{Segmentation ROC curves and AUC for each tested region.}
    \label{fig:mpi_seg_curves}
\end{figure*}

Figures \ref{fig:mpi_det_curves} and \ref{fig:mpi_seg_curves}, and Table \ref{table:results_mpi} show the results obtained by the tested techniques for each region.
In Table \ref{table:results_mpi} the bold text indicates the best results for each metric.
The results show that the proposed method outperforms or has similar results compared to approaches that attain state-of-art results in the MVTec-AD dataset.

For simple anomaly detection (measured by the detection ROC-AUC, see Figure \ref {fig:mpi_det_curves}), our method, PaDiM and SPTM present similar performance in most cases.
The proposed method shows detection ROC-AUC 1.0 in 4 out of 6 regions, which means it identified all modifications in these regions.
SPADE and DFR presented significantly worse results.
This is explained by the difficulty of finding a threshold that attains a good trade-off between TPR and FPR.

As for anomaly segmentation (Figure \ref {fig:mpi_seg_curves}), all the methods achieved ROC-AUC higher than 0.9 for almost every region.
That shows these methods can segment most of the anomalies correctly.
The proposed method and PaDiM showed the best average performance.
The difference between detection and segmentation ROC-AUC results for SPADE and DFR is explained by the class imbalance in each problem.
For detection, the test set is balanced, since it contains the same number of positive and negative samples. 
However, for per-pixel segmentation, the classes are very imbalanced, with less than 2\% being positive pixels.
This allows the model to generate small segmentation errors in several images without impacting segmentation ROC-AUC, but with high impact in detection ROC-AUC.

\begin{figure*}[!ht]
    \centering
    \captionsetup[subfigure]{labelformat=empty}
    \begin{subfigure}{.12\linewidth}
        \begin{subfigure}{\linewidth}
            \centering
            \includegraphics[width=\linewidth]{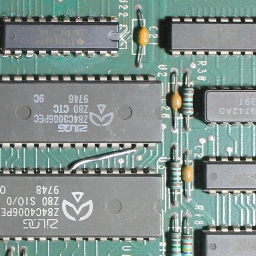}
        \end{subfigure}
        \begin{subfigure}{\linewidth}
            \centering
            \includegraphics[width=\linewidth]{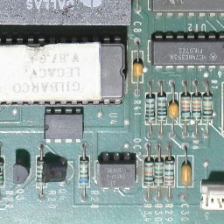}
        \end{subfigure}
        \begin{subfigure}{\linewidth}
            \centering
            \includegraphics[width=\linewidth]{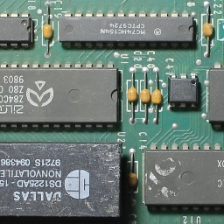}
        \end{subfigure}
        \begin{subfigure}{\linewidth}
            \centering
            \includegraphics[width=\linewidth]{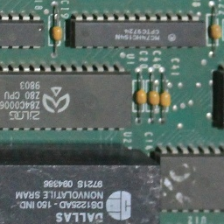}
        \end{subfigure}
        \begin{subfigure}{\linewidth}
            \centering
            \includegraphics[width=\linewidth]{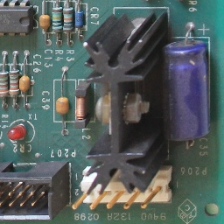}
        \end{subfigure}
        \caption{Input image}
    \end{subfigure}
    \begin{subfigure}{.12\linewidth}
        \begin{subfigure}{\linewidth}
            \centering
            \includegraphics[width=\linewidth]{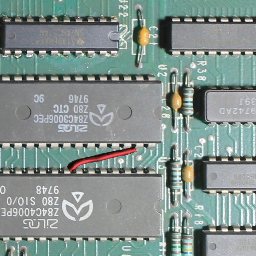}
        \end{subfigure}
        \begin{subfigure}{\linewidth}
            \centering
            \includegraphics[width=\linewidth]{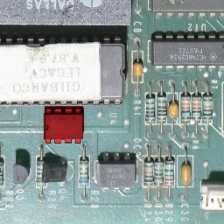}
        \end{subfigure}
        \begin{subfigure}{\linewidth}
            \centering
            \includegraphics[width=\linewidth]{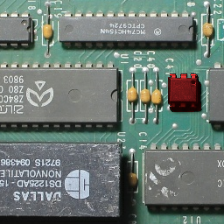}
        \end{subfigure}
        \begin{subfigure}{\linewidth}
            \centering
            \includegraphics[width=\linewidth]{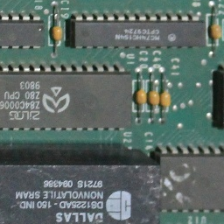}
        \end{subfigure}
        \begin{subfigure}{\linewidth}
            \centering
            \includegraphics[width=\linewidth]{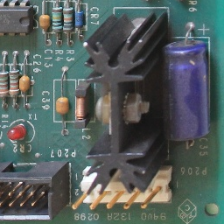}
        \end{subfigure}
        \caption{Ground truth}
    \end{subfigure}
    \begin{subfigure}{.12\linewidth}
        \begin{subfigure}{\linewidth}
            \centering
            \includegraphics[width=\linewidth]{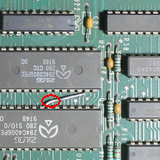}
        \end{subfigure}
        \begin{subfigure}{\linewidth}
            \centering
            \includegraphics[width=\linewidth]{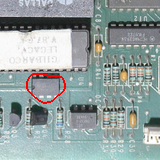}
        \end{subfigure}
        \begin{subfigure}{\linewidth}
            \centering
            \includegraphics[width=\linewidth]{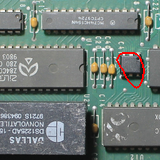}
        \end{subfigure}
        \begin{subfigure}{\linewidth}
            \centering
            \includegraphics[width=\linewidth]{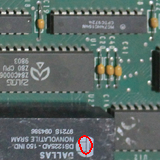}
        \end{subfigure}
        \begin{subfigure}{\linewidth}
            \centering
            \includegraphics[width=\linewidth]{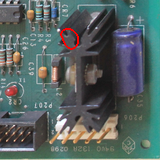}
        \end{subfigure}
        \caption{PaDiM}
    \end{subfigure}
    \begin{subfigure}{.12\linewidth}
        \begin{subfigure}{\linewidth}
            \centering
            \includegraphics[width=\linewidth]{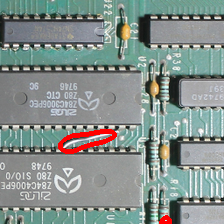}
        \end{subfigure}
        \begin{subfigure}{\linewidth}
            \centering
            \includegraphics[width=\linewidth]{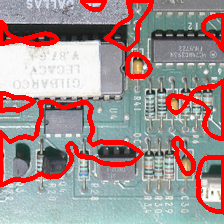}
        \end{subfigure}
        \begin{subfigure}{\linewidth}
            \centering
            \includegraphics[width=\linewidth]{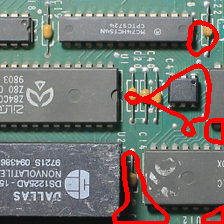}
        \end{subfigure}
        \begin{subfigure}{\linewidth}
            \centering
            \includegraphics[width=\linewidth]{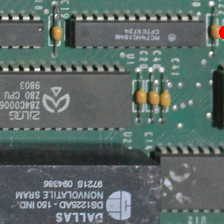}
        \end{subfigure}
        \begin{subfigure}{\linewidth}
            \centering
            \includegraphics[width=\linewidth]{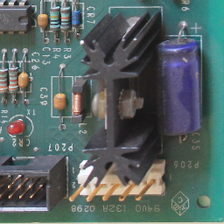}
        \end{subfigure}
        \caption{SPADE}
    \end{subfigure}
    \begin{subfigure}{.12\linewidth}
        \begin{subfigure}{\linewidth}
            \centering
            \includegraphics[width=\linewidth]{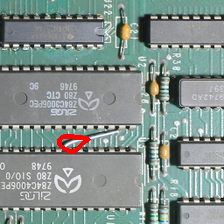}
        \end{subfigure}
        \begin{subfigure}{\linewidth}
            \centering
            \includegraphics[width=\linewidth]{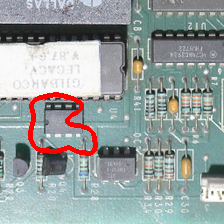}
        \end{subfigure}
        \begin{subfigure}{\linewidth}
            \centering
            \includegraphics[width=\linewidth]{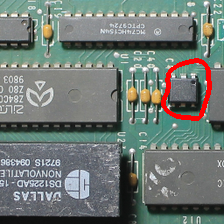}
        \end{subfigure}
        \begin{subfigure}{\linewidth}
            \centering
            \includegraphics[width=\linewidth]{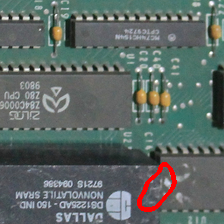}
        \end{subfigure}
        \begin{subfigure}{\linewidth}
            \centering
            \includegraphics[width=\linewidth]{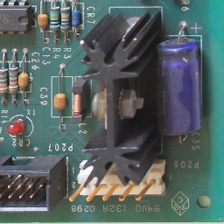}
        \end{subfigure}
        \caption{SPTM}
    \end{subfigure}
    \begin{subfigure}{.12\linewidth}
        \begin{subfigure}{\linewidth}
            \centering
            \includegraphics[width=\linewidth]{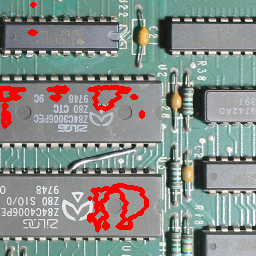}
        \end{subfigure}
        \begin{subfigure}{\linewidth}
            \centering
            \includegraphics[width=\linewidth]{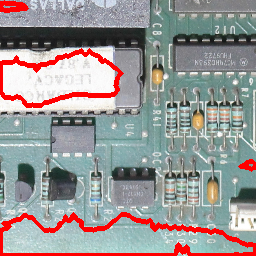}
        \end{subfigure}
        \begin{subfigure}{\linewidth}
            \centering
            \includegraphics[width=\linewidth]{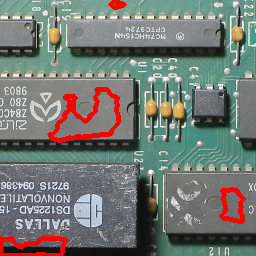}
        \end{subfigure}
        \begin{subfigure}{\linewidth}
            \centering
            \includegraphics[width=\linewidth]{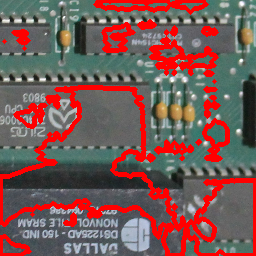}
        \end{subfigure}
        \begin{subfigure}{\linewidth}
            \centering
            \includegraphics[width=\linewidth]{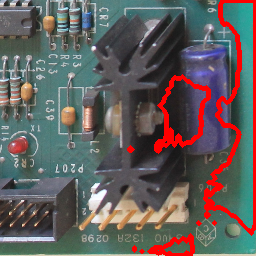}
        \end{subfigure}
        \caption{DFR}
    \end{subfigure}
    \begin{subfigure}{.12\linewidth}
        \begin{subfigure}{\linewidth}
            \centering
            \includegraphics[width=\linewidth]{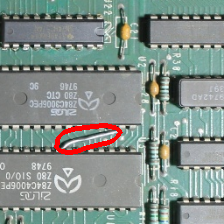}
        \end{subfigure}
        \begin{subfigure}{\linewidth}
            \centering
            \includegraphics[width=\linewidth]{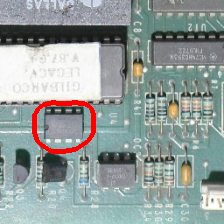}
        \end{subfigure}
        \begin{subfigure}{\linewidth}
            \centering
            \includegraphics[width=\linewidth]{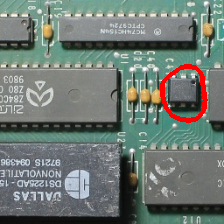}
        \end{subfigure}
        \begin{subfigure}{\linewidth}
            \centering
            \includegraphics[width=\linewidth]{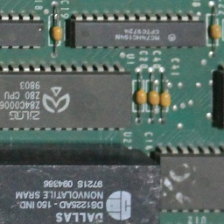}
        \end{subfigure}
        \begin{subfigure}{\linewidth}
            \centering
            \includegraphics[width=\linewidth]{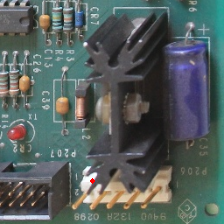}
        \end{subfigure}
        \caption{Ours}
    \end{subfigure}
    
    \caption{Segmentation comparison between our method and the baseline methods in image regions grid3\_2, grid2\_3 and grid2\_2 with modifications, and in regions grid2\_2 and grid4\_3 without modifications. The red contours represents the anomalies detected by the evaluated algorithms.
    }
    \label{fig:comparasion_iou}
\end{figure*}

Despite the similar ROC-AUC results obtained by our approach and PaDiM, we observed that the segmentation in several samples was visibly different.
We noticed that this happened because of the imbalance between positive and negative pixels, which leads to high ROC-AUC values even when the model produces false positive classifications.
IoU can express the segmentation precision better than the ROC-AUC, being more sensitive to incorrectly classified pixels and, consequently, to deviations in the shape and size of the segmented objects.
This can be seen in Figure~\ref{fig:comparasion_iou}, which shows some segmentation samples produced by our technique and by the baseline methods.
We note that most baseline methods had several false positives, i.e.~these methods successfully localize the modifications in a general manner, but several additional pixels are detected, so the segmented shape does not match the anomaly.
Generally, the models identify large regions around modifications or smaller shapes which do not cover an entire component.
That might be interpreted as a false detection by a human inspector without specialized knowledge, because it does not cover just a component but a region that includes parts of other components.
Additionally, Figure \ref{fig:comparasion_iou} shows that some of the baseline methods can produce more false positive detections when there are no anomalies in the board.

Regarding the IoU, the proposed method outperformed the baseline methods for all evaluated regions, achieving an IoU higher than 0.5 for all regions --- this is a relevant mark, since challenges such as Pascal VOC\footnote {http://host.robots.ox.ac.uk/pascal/VOC/index.html} and COCO use IoU $>$ 0.5 as one possible criterion for a successful detection.
Note that the IoU is sensitive to the size of the modification, as the weight of an incorrectly classified pixel is higher for smaller objects.
Our method was able to segment small modifications, like the jumper wire in the ``grid3\_2'' region (the first row in Fig.~\ref{fig:comparasion_iou}).
PaDiM presented IoU close to 0.5 for all regions, except for the ``grid3\_2'' region, which contains the smallest modification --- there was a high number of false negatives, which led to a partial segmentation.
As for SPADE, SPTM, and DFR, performance was worse in several cases.
As discussed above, these techniques displayed a higher number of false positives, segmenting large regions around the modifications, and in many cases detecting modifications where none exist.
That can be explained by the lighting and perspective variations in this dataset.

The difference between the segmentation quality of the techniques is reinforced if we observe the precision, recall and F-score metrics.
Our method presented the best segmentation precision for all regions, meaning that it could better detect pixels that represent anomalies with fewer false positives.
At the same time, regarding segmentation recall, our method outperforms the baseline methods for almost all regions by a significant margin, showing that the proposed method presents less false negatives.
Our method's advantages are reflected in the average F-score, which is significantly higher than the one achieved by the baseline methods.

In conclusion, the obtained results show that, while all techniques are able to detect and segment modifications (as indicated by the detection and segmentation ROC-AUC metrics), the proposed method can better approximate the shape of objects (as indicated by IoU, precision, recall and F-score). In a practical scenario, this advantage can help a human inspector identify the specific components that characterize a modification.


\subsection{Results on the MVTec-AD dataset}

\begin{figure*}[!ht]
    \centering
    \captionsetup[subfigure]{labelformat=empty}
    \begin{subfigure}{.48\linewidth}
        \centering
         \includegraphics[width=\linewidth]{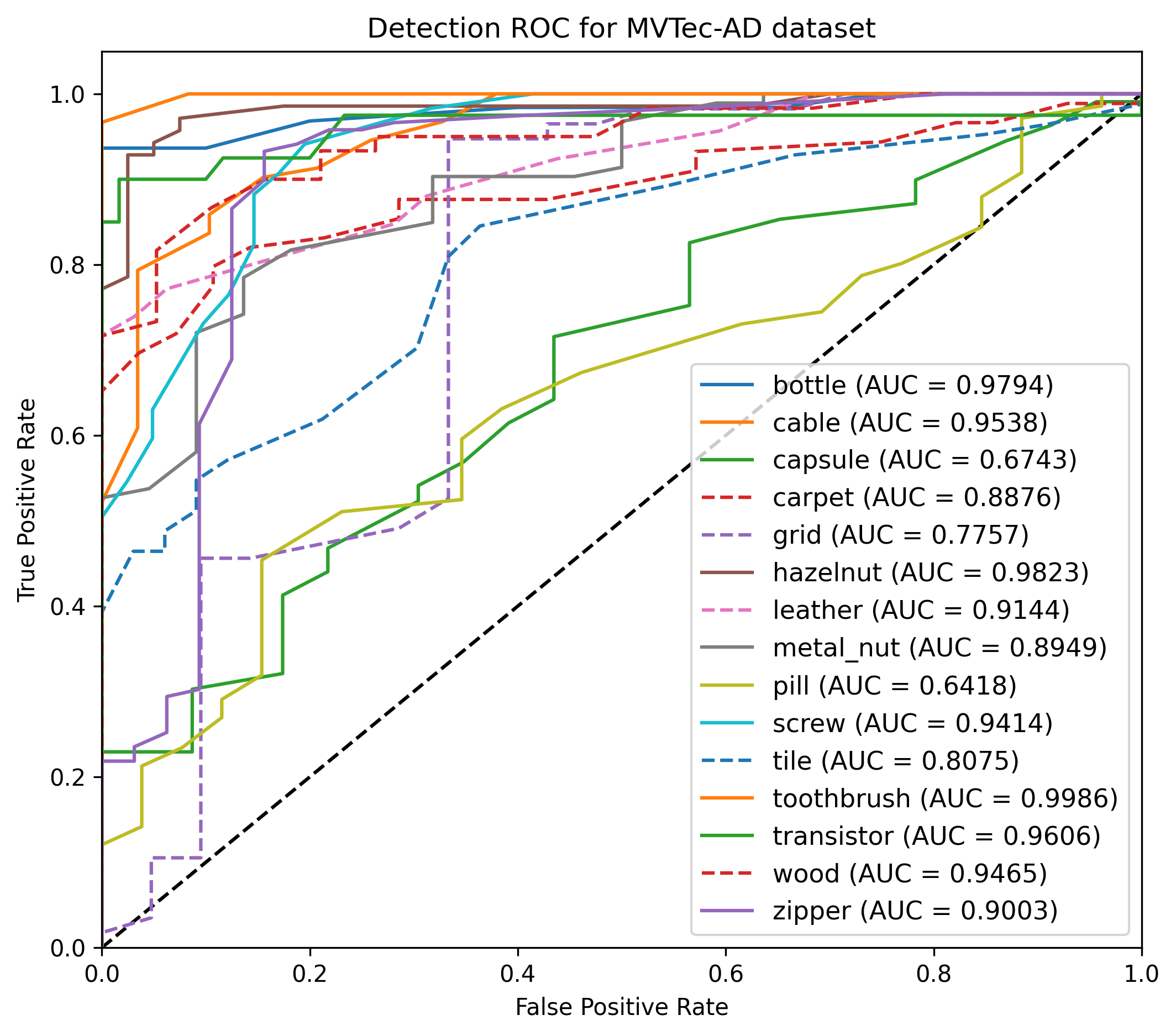}
        \caption{}
    \end{subfigure}
    \begin{subfigure}{.48\linewidth}
        \centering
         \includegraphics[width=\linewidth]{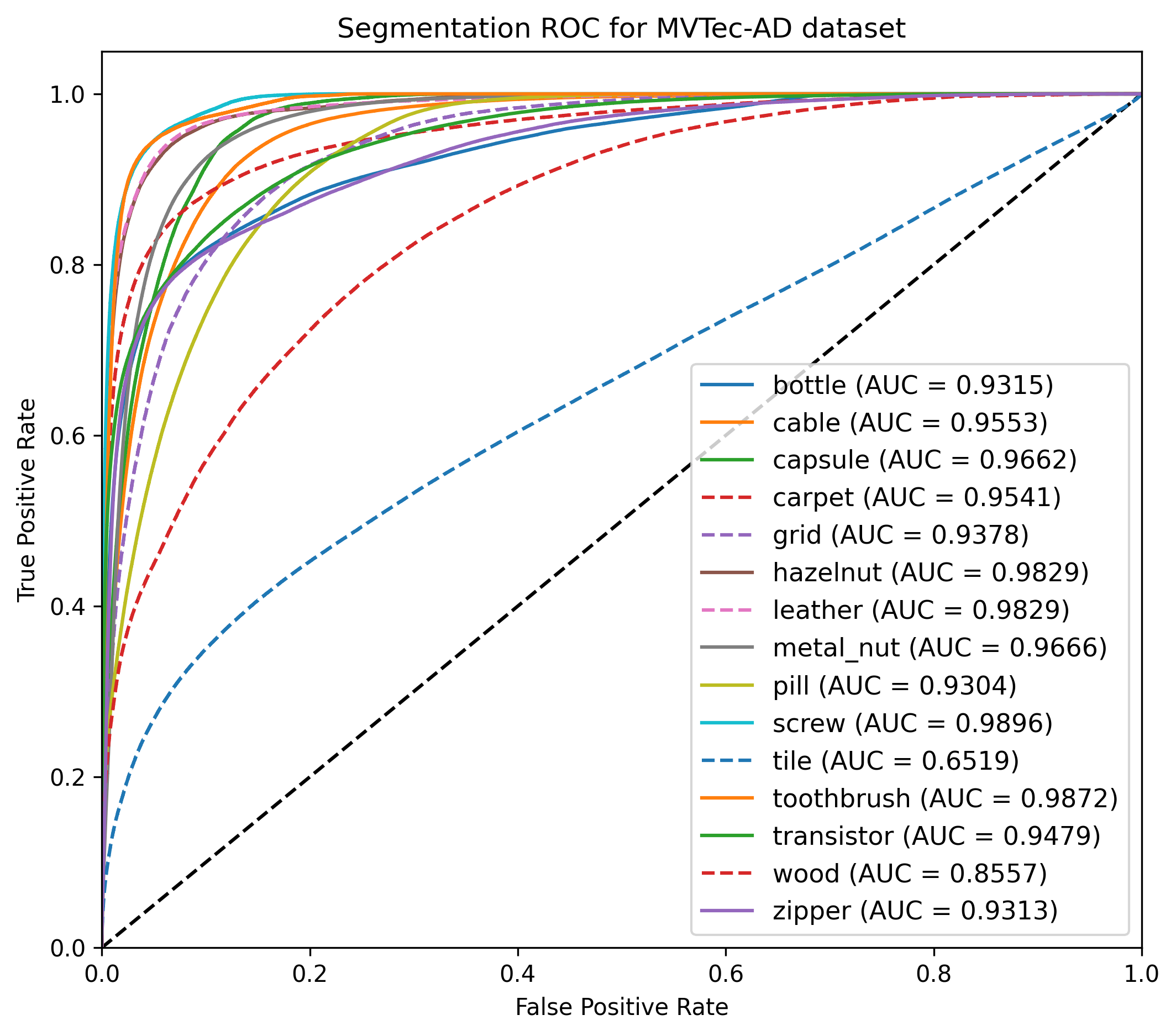}
        \caption{}
    \end{subfigure}
   
    \caption{Detection and segmentation ROC and AUC of our method for all textures and objects in the MVTec-AD dataset. Solid lines are used for the ``object'' category and dashed for the ``textures'' category.}
    \label{fig:mvtec_curves}
\end{figure*}

To evaluate the performance of our method for other anomaly localization contexts, apart from the PCB modifications it was designed for, we tested it along with the baseline methods on the MVTec-AD dataset.
Figure~\ref{fig:mvtec_curves} shows the detection and segmentation ROC for all objects and textures from the dataset.
Table~\ref{table:results_mvtec} shows the evaluated metrics following the categorization defined by \citep{Bergmann}, with anomalies grouped by type: ``objects'' and ``textures''.
The former shows certain types of objects, with most anomalies involving the addition, removal, or modification of parts or components; while the latter shows close-ups of surfaces, with anomalies consisting of alterations to a common texture pattern.

\begin{table}[!htb]
\small
\centering
\caption{Results of the proposed method and the baseline methods for the MVTec-AD dataset. We show the results for two main categories defined by: textures and objects.}
\label{table:results_mvtec}
\begin{tabular}{lllll}
\hline
\multicolumn{1}{c}{Metric}        & \multicolumn{1}{c}{Method} & \multicolumn{1}{c}{Tex.} & \multicolumn{1}{c}{Obj.}  \\ \hline
\multirow{5}{*}{\shortstack[l]{Detection \\ ROC-AUC}}    & Ours                          & 0.870                        & 0.890  \\
                                  & PaDiM                         & 0.960                        & 0.880 \\
                                  & SPADE                         & 0.860                        & 0.850 \\
                                  & DFR                           & 0.930                        & 0.910 \\
                                  & SPTM                          & \textbf{0.980}                        & \textbf{0.930} \\ \hline
\multirow{5}{*}{\shortstack[l]{Segmentation \\ ROC-AUC}} & Ours                          & 0.880                        & 0.960                       \\
                                  & PaDiM                         & 0.950                        & \textbf{0.970} \\
                                  & SPADE                         & \textbf{0.970}                        & 0.960 \\
                                  & DFR                           & 0.910                        & 0.940 \\
                                  & SPTM                          & 0.960                        & 0.870 \\ \hline
\multirow{5}{*}{IoU}              & Ours                          & 0.290                        & \textbf{0.430} \\
                                  & PaDiM                         & 0.330                        & 0.410 \\
                                  & SPADE                         & \textbf{0.380}                        & 0.420 \\
                                  & DFR                           & 0.310                        & 0.310 \\
                                  & SPTM                          & 0.320                        & 0.380 \\ \hline
\multirow{5}{*}{\shortstack[l]{Segmentation \\ Precision}}        & Ours                          & 0.440                        & \textbf{0.562} \\
                                  & PaDiM                         & 0.408                        & 0.485 \\
                                  & SPADE                         & \textbf{0.460}                        & 0.518 \\
                                  & DFR                           & 0.364                        & 0.481 \\
                                  & SPTM                          & 0.364                        & 0.389 \\ \hline
								  
\multirow{5}{*}{\shortstack[l]{Segmentation \\ Recall}}           & Ours                          & 0.458                        & 0.625 \\
                                  & PaDiM                         & 0.628                        & \textbf{0.677} \\
                                  & SPADE                         & \textbf{0.682}                        & 0.640 \\
                                  & DFR                           & 0.614                        & 0.515 \\
                                  & SPTM                          & 0.598                        & 0.575 \\ \hline
								  
\multirow{5}{*}{\shortstack[l]{Segmentation \\ F-score}}         & Ours                          & 0.446                        & \textbf{0.591} \\
                                  & PaDiM                         & 0.490                        & 0.560 \\
                                  & SPADE                         & \textbf{0.546}                        & 0.571 \\
                                  & DFR                           & 0.452                        & 0.478 \\
                                  & SPTM                          & 0.448                        & 0.382 \\ \hline
\end{tabular}
\end{table}

According to Table \ref {table:results_mvtec}, our method did not perform as well as the baseline methods for the ``texture'' category.
This behavior can be explained by the way we combined the content loss function with the pixel-wise mean squared error.
In other tasks, content loss is usually employed in conjunction with the ``style loss'' function, which tries to keep feature distributions in each layer the same in both the image and its reconstruction.
Content loss only captures the aspect of image structures, while MSE compares individual pixels in the image and its reconstruction.
That means our model is less capable of representing general texture patterns, being directed towards representing structures and pixel organizations observed during training (on the other hand, that is what allows our approach to detect even small anomalies).
The problem is exacerbated by the small number of training samples in the MVTec-AD dtaset, which has only around 50 training images per class.

As for the ``objects'' category, the proposed method performed similarly to the baseline methods, particularly SPADE and PaDiM.
That indicates our method may present better results for problems where most anomalies or modifications are the addition or removal of objects in the inspected area. 


\subsection{Discussion}

The results show that our method can successfully segment anomalies in images of assembled PCBs taken without tight control over perspective and illumination conditions.
In the MPI-PCB dataset, our method outperformed the state-of-the-art baseline methods, showing superior performance on segmentation and detection.
For anomaly segmentation, our method presented the approximated shape of the anomalies in all evaluated regions, showing less false positive and false negative pixels.
A better segmentation may be advantageous for a human inspector identifying specific components as modifications.
The experiments performed on the MVTec-AD dataset demonstrated that our method can be used for anomaly detection in other contexts, when the analyzed object or surface does not contain textures with random patterns.

One limitation of the proposed approach is that it is only capable of detecting modifications that are visible in the images and that form structures occupying groups of pixels --- that means it may fail if the images have very poor quality or low resolution.
Although this can be avoided simply by using good cameras and taking some care when capturing the images, invisible modifications are still undetectable --- e.g.~some modifications are hidden below a chip, which is removed and resoldered; and others involve replacing memory units or cloning components.
These modifications cannot be detected by any vision-based approach, requiring radically different approaches, such as electrical tests or completely disassembling the board.
However, we note that our approach was mainly designed to support the work of human inspectors, which in the considered scenario perform their work based solely on visual cues, so detecting this kind of invisible modification is outside the scope of our work.

%% file: 6_conclusion.tex
\section{Conclusions}
\label{section.conclusion}

In this paper, we addressed the problem of detecting modifications in PCBs based on photographs.
For that purpose, we proposed a reconstruction-based anomaly detection method using a CAE architecture, trained using just anomaly-free samples with a combination of the content loss and the mean squared error functions.
We also introduced MPI-PCB, a labeled PCB image dataset for training and evaluating anomaly detection and segmentation methods.
Experiments on that dataset showed that our method has superior results for modification segmentation when compared to other state-of-art methods.
We also performed experiments in the popular MvTec-AD dataset, with our method attaining results close to other methods when detecting anomalies such as adding or removing objects, showing that it can be employed in other problem domains.

In future research, we plan to create a more varied dataset, with a greater number of modifications to evaluate the performance in other situations, such as very small modifications.
Another possible improvement is designing a loss function capable of better learning texture information, based on techniques such as adversarial learning.